%% file: twitter_paper-mo.tex
\documentclass[journal]{IEEEtran}

\usepackage{setspace} 

\ifCLASSINFOpdf
  \usepackage[pdftex]{graphicx}
  \DeclareGraphicsExtensions{.pdf,.jpeg,.png}
\else
\fi
\begin{document}
%
\title{Leveraging Twitter for Low-Resource Conversational Speech Language Modeling}
%
%
%

\author{Aaron Jaech,
        and Mari Ostendorf
\thanks{Supported by the Intelligence Advanced Research Projects Activity (IARPA) via Department of Defense US Army Research Laboratory contract number W911NF-12-C-0014. The U.S. Government is authorized to reproduce and distribute reprints for Governmental purposes notwithstanding any copyright annotation thereon. Disclaimer: The views and conclusions contained herein are those of the authors and should not be interpreted as necessarily representing the official policies or endorsements, either expressed or implied, of IARPA, DoD/ARL, or the U.S. Government.}
\thanks{A. Jaech and M. Ostendorf are with the Department
of Electrical Engineering, University of Washington, Seattle, WA 98195
USA e-mail: \{ajaech,ostendor\}@uw.edu.}
}

\maketitle

\begin{abstract}
In applications involving conversational speech, data sparsity is
a limiting factor in building a better language model. We propose a simple, language-independent method
 to quickly harvest large amounts of data from Twitter to supplement a smaller training set that is more closely
matched to the domain. The techniques lead to a significant reduction in perplexity on four low-resource languages even though the
presence on Twitter of these languages is relatively small. 
We also find that the Twitter text
is more useful for learning word classes than the in-domain text and that use of these word classes leads to 
further reductions in perplexity. Additionally, we introduce a method of using social and textual information
to prioritize the download queue during the Twitter crawling. This maximizes the amount of useful data that
can be collected, impacting both perplexity and vocabulary coverage.
\end{abstract}

%
\IEEEpeerreviewmaketitle

\section{Introduction}
%
%
%
%

\input{intro2}


\section{Related Work}
\label{section:related}

\input{related2}

\section{Twitter Text for Language Modeling}
\label{section:twitter}

\subsection{Twitter Data Collection}
\label{section:methodology}
A small set of in-domain data forms the basis from which the data collection is
bootstrapped. We want to query Twitter using phrases from the in-domain data to find additional
phrases from a similar distribution. The queries are
composed of bigrams from the in-domain data taken in descending order of frequency. Not all 
bigrams from the in-domain data are suitable queries. For example, due to code switching
the randomly chosen query may not actually be in the target language. If the chosen query
happens to be in English but the target language is Bengali, then keeping this query could
pose a significant problem. If the query matches an English bigram then an overwhelming amount
of irrelevant results will be returned. This can be avoided by checking that the query results
overlap in vocabulary with the in-domain data and discarding queries that fail to meet a minimum
threshold. In our experiments, we required an in-vocabulary hit rate of at least 10\%.

Instead of using the query results directly, we take the set of users who matched our query and then
collect the entire history (up to 2,000 tweets per user) of those users' posts. By taking the users' 
histories, instead of using the query results directly, we gather sentences that have 
no overlap with our query words and avoid over-representing the queries in the collected data.

Like previous researchers \cite{web20}, we find it necessary to do significant
cleaning of the Twitter text before building any language models.
In the cleaning process, tokens such as URLs and hashtags are removed but other out-of-vocabulary (OOV)
tokens such as mentions (``@username'') are left in place to provide appropriate context breaks
when building the language model. Capitalization is standardized when appropriate for that language.
We perform duplicate sentence removal to deal with spam and retweets. We found a consistent
but small gain from exempting very short tweets (those consisting of one or two words) from the
de-duplication. Presumably, this is because many short phrases are naturally repeated many times
in conversation but the same is not true for longer phrases. 

After the text is cleaned, much of the collected data is still not useful for training the 
language model due to topic and style mismatch. We would like to select, from all of the 
collected text, the sentences that
are the best match to our target domain. 

When the in-domain data is very limited, 
in-vocabulary hit rate with
respect to the target domain provides a useful proxy for stylistic and topical similarity. In prior work \cite{Bulyko07}, we found that a two-stage process of vocabulary hit rate followed by perplexity filtering was useful.  
Here, efforts to use the estimated n-gram probabilities to rank the Twitter sentences were not
successful. There are two possible reasons for this.  First, the in-domain LM training data is small and perplexity estimates
on individual sentences have high variance. Second, for the high vocabulary growth rate languages explored here,
many of the words are not in-vocabulary, so n-gram context is often broken and unigrams dominate the score.

\begin{figure}[!t]
 \centering
 \includegraphics[scale=0.5]{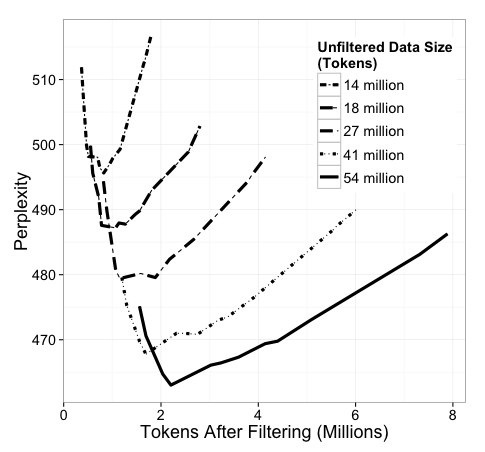}
 \caption{Optimal IV Threshold for Filtering Bengali Twitter Data}
 \label{fig:iv_threshold}
\end{figure}

We put a threshold on the in-vocabulary (IV) hit rate to decide which sentences we will keep and choose the 
threshold that minimizes perplexity on a held-out set.  Obviously, there is a trade-off between
the amount of data used to build the language model and the degree to which the data matches the target domain.
The optimal threshold depends upon the amount of data we have to begin with. As the size of the
unfiltered data increases, the optimal filtering threshold becomes more aggressive. This is most easily seen
in Figure \ref{fig:iv_threshold} when the unfiltered data size ranges from 14 million to 27 million tokens.
Even though the 27 million token data is almost twice as big as the 14 million token one, the more aggressive 
filtering threshold on the bigger source causes them both to have about the same number of tokens after filtering.
In our experiments the optimal vocabulary hit rate threshold tends to fall between $50\%$ and $70\%$. Typically, only a small
percentage of sentences have IV hit rates above the chosen threshold.

\subsection{Language Modeling Approaches}
\label{section:modelling}

After filtering, the Twitter LM is built from the remaining sentences concatenated with the in-domain data.
The LM is a trigram model with modified Kneser-Ney smoothing and is built using the SRILM toolkit \cite{stolcke02}. 
This is interpolated with another LM built only from the target domain data. Here, the vocabulary is restricted
to the words found in the in-domain data.

Concatenating the Twitter data with the in-domain data  
ensures that the full vocabulary from the in-domain data is represented in the language model
training data. Thus, the mixture model combines two components: in-domain only and the in-domain plus
Twitter combination. The size of the in-domain data, in our case, is typically less than 1\% of the size
of the Twitter data. Except for rare words that only appear in the in-domain data, the n-gram counts of
the in-domain plus Twitter combination are dominated by the Twitter data.

As noted earlier, prior work has shown that class-based models can be useful for combining data from multiple domains, though much of the work used part-of-speech (POS) classes.
There is no POS tagger readily 
available for the languages we are working with. Instead, we need to learn the word class
assignments from the data. Specifically, we use the unsupervised hierarchical clustering approach from Brown et al. \cite{brown92}. The
clustering algorithm attempts to find the assignment of words to clusters that maximizes the mutual
information between adjacent tokens in the data. This corresponds to maximizing the likelihood
of the data assuming a bigram model.

The benefit of using word classes is that the resulting language model can have fewer parameters
than a word-based n-gram model. In our case, the in-domain training text is so small that
it can be a real advantage to have fewer parameters to learn in the language model. The problem is
that if the data is too small to reliably estimate word transition probabilities ($p(w_i | w_{i-1})$),
then it will also be difficult to learn a good partitioning of the words into classes. 

Our hypothesis is that the advantage in learning the word class assignments on the Twitter data, 
which solves the data sparsity problem,
outweighs any performance penalty that is incurred due to domain mis-match. This differs from the traditional
approach where the class assignments ($w_i \in c_j$) are learned from the same
training text which is used to estimate the class transition probabilities ($p(c_i|c_{i-1})$) and the 
word probabilities ($p(w_i | c_j)$). Twitter data, in these experiments, refers to the concatenation of the
text downloaded from Twitter with our in-domain data for the reasons described above. The experiments
in Section \ref{section:class_experiments} will compare learning the class assignments on out-of-domain
data (hybrid method) to the traditional approach (baseline).

\section{Experiments}
\label{section:experiments}

\subsection{Experimental Data}
\label{section:babel}
The in-domain data used in the experiments in this paper comes from the IARPA Babel
program.\footnote{http://www.iarpa.gov/index.php/research-programs/babel}
This program focuses on keyword search for low-resource languages. The languages are low
resource in the sense that they have fewer native speakers than the languages receiving the
most attention from researchers and also in the sense that the provided training data is
small in comparison to what is typically used. We exclusively focused on the so-called 
limited language pack, which consists of only ten hours of recorded telephone conversations.
Our experiments were conducted using the Bengali, Tamil, Turkish and Zulu languages. The 
languages that we selected for our experiments have the largest vocabulary sizes (See \mbox{Table \ref{babel}}) 
of the  languages in the Babel program and thus suffer the most from the data sparsity problem.
 As a point of comparison, Tagalog, which was not used in our experiments, has one third as many vocabulary items as Tamil. 

\begin{table}[h]
\centering
\caption{Language Vocabulary Size}
\begin{tabular}{|crr|}\hline
\textbf{Language} & \multicolumn{1}{c}{\textbf{Types}} & \multicolumn{1}{c|}{\textbf{Tokens}} \\ \hline
Bengali & 7,932 & 72,614 \\
Tamil & 14,264 & 70,258 \\
Turkish & 10,069 & 67,362 \\
Zulu & 13,628 & 58,027 \\ \hline
\end{tabular}
\label{babel}
\end{table}

\subsection{N-gram Language Models}
\label{section:ngram}

We performed data collection and language model training experiments on the Bengali, Tamil, Turkish and
Zulu languages. We were able to collect useful data for each of the four languages as seen in 
\mbox{Table \ref{baseline_improvements}}. The data collection experiment was especially successful for Turkish 
and Bengali. For both of those languages the interpolation weight given to the Twitter LM was over 20\% and
the corresponding reduction in perplexity was more than 12\%.

\begin{table}[h]
\centering
\caption{Twitter Data Collection Experiment Results}
\begin{tabular}{|crrrc|} \hline
\textbf{Lang.} & \textbf{Users} & \textbf{Lines} & \textbf{$\Delta$ PPL} & \textbf{Weight} \\ \hline
Bengali    & 12k   & 7.7M   & 12.5\%  & .21   \\
Tamil       & 13k  & 4.7M   & 7.5\%   & .18 \\
Turkish   & 7k     & 13.6M  & 13.6\%  & .27 \\
Zulu       & 3k    & 5.2M   & 5.1\%   & .11 \\ \hline
\end{tabular}
\label{baseline_improvements}
\end{table}

A large amount of data was also collected for Zulu and Tamil although the perplexity reduction was not
as large as it was for Bengali and Turkish. There are a few possible explanations for this outcome. Both
Zulu and Tamil have larger vocabularies (See Table \ref{babel}) than the other languages, which makes it
hard to find enough webtext to adequately cover the vocabulary. For Zulu, there is an overlap in the character
set with English and many Zulu speakers also tweet in English. The official Babel data also has some English
words in the Zulu vocabulary. These factors combine to make it difficult to prevent the Zulu Twitter data from
being polluted by a large amount of English text and could explain the relatively small gain for Zulu.  However, as we shall see in the next section, the Zulu data is quite useful in the class n-gram model.

\subsection{Class N-gram Models}
\label{section:class_experiments}

Experiments were performed for the Bengali, Tamil, Turkish, and Zulu languages. We modified Liang's \cite{liang} 
implementation of the Brown clustering algorithm \cite{brown92} for our experiments. Our modifications
take care of merging the class assignments from the Twitter data with the probability estimates from the
in-domain data. Another modification used multithreading to increase the speed of the algorithm, making
it tractable to operate on larger vocabularies. Each clustering experiment uses 500 word classes. The class
LMs are trigram models with Witten-Bell smoothing \cite{witten1991zero}.

The hybrid method differs from the baseline in the use of one set of data (from Twitter) to learn the class
assignments and another set of data to learn the word and class probabilities (Babel data). We used two
baselines to compare our hybrid method with the traditional approach as applied to both the Twitter data
and the Babel data respectively. 
As before, the Twitter data for these experiments has been concatenated with the in-domain data.

\begin{table}[h]
\centering
\caption{Perplexity of Word Class Language Models}
\begin{tabular}{|cccc|}\hline
\textbf{Language} & \textbf{\begin{tabular}[c]{@{}c@{}}In-Domain\\ Baseline\end{tabular}} & \textbf{\begin{tabular}[c]{@{}c@{}}Twitter\\ Baseline\end{tabular}} & \textbf{\begin{tabular}[c]{@{}c@{}}Hybrid\\ Method\end{tabular}} \\ \hline
Bengali & 284 & 314 & 256 \\
Tamil & 441 & 397 & 367 \\ 
Turkish & 280 & 263 & 236 \\
Zulu & 318 & 286 & 240 \\ \hline
\end{tabular}
\label{baseline_comparison}
\end{table}

In Table \ref{baseline_comparison} we list the perplexity of the class language models on the held-out
in-domain data. For each of the three languages, the hybrid method has a substantially lower perplexity
than either of the baselines. The average perplexity improvement is more than 10\%. This validates our 
hypothesis that the word classes learned from the Twitter data would generalize well.

\begin{table}[h]
\centering
\caption{Turkish LM Interpolation Results}
\label{my-label}
\begin{tabular}{|ccc|} \hline
\textbf{Model} & \textbf{Babel} & \textbf{Babel + Twitter} \\ \hline
Word Trigram & 245 & 210 \\
Class Trigram & 280 & 236 \\
Word + Class & 239 & 199 \\ \hline
\end{tabular}
\label{lm_interpolation}
\end{table}

It is also important to look at what happens when we interpolate the class LM with the word
based LMs from Section \ref{section:ngram}. Specifically we were interested in finding whether
there is an advantage in interpolating the class
n-gram model with our word-based language models and how much
each language model contributes to the result. Table \ref{lm_interpolation} lists the
perplexities of each of the individual LMs and the combination LM for Turkish. Although the class-based
LM by itself is worse than the Twitter word-based LM, the word + class combination LM is substantially 
better than any of the individual models. In the interpolated combination, the class LM
received 38\% of the weight and the Babel + Twitter word LM received 62\% of the weight.  Using the same technique
of combining a word LM with a class LM without the Twitter data gives only a small 3\% reduction in perplexity. 
The 19\% reduction in perplexity over the Babel word trigram model, obtained by the Twitter word+class combination
LM, comes from the superior class assignments learned on the Twitter data.

\begin{table}[h]
\centering
\caption{Perplexity Improvements with Class Based LMs}
\begin{tabular}{|rccc|}\hline
\multicolumn{1}{|c}{\textbf{Language}} & \textbf{Baseline} & \textbf{Combined} & \textbf{$\Delta$ PPL} \\
\hline
Bengali & 268 & 222 & 17\% \\
Tamil & 381 & 325 & 15\% \\
Turkish & 245 & 199 & 19\% \\
Zulu &  289 & 234  &    19\% \\ \hline
\end{tabular}
\label{perplexity_improvement}
\end{table}

Similar perplexity reductions were achieved by the combination LMs for Bengali, Tamil, and Zulu. (See
\mbox{Table \ref{perplexity_improvement}}.) Tamil, whose 15\% perplexity improvement is the least of all the languages,
is still twice the improvement obtained in the initial experiment from Table \ref{baseline_improvements}.
For Zulu, the initial experiments failed to give much of an improvement in perplexity but using the class LMs
a strong perplexity reduction of 19\% was obtained. This is in spite of the relatively small amount of Twitter data
available in these two languages.

\section{Using Social Information to Prioritize Data Collection}
\label{section:crawling}

Due to the strong relationship between the amount of available Twitter data
and the LM perplexity (see Figure \ref{fig:iv_threshold}), it is desirable to have as
much useful data as possible.
After obtaining an initial download of Twitter data in response to the selected queries,
it is possible to expand our search across social connections. Adding the social connections
of the already downloaded users to the download queue 
will allow us to discover new content when our source of seed queries is
exhausted. However, any attempt at crawling the Internet will quickly run into the problem that 
the rate of growth of the download queue far exceeds the rate at which
it can be processed. For example, after processing the first 4,000
Tamil speaking Twitter users there were 200,000 additional users in the queue.
Internet search engines deal with this problem by prioritizing the queue such
that the most important or useful pages are visited first \cite{crawling}. If 
such a strategy were not employed, many useful pages would be so far back in the
queue that they would effectively never be reached. Such concerns are even more relevant
when the data is accessed through an API that places limits or costs on each request.
In this section, we look at the feasibility of applying queue ranking methods to our task.

\subsection{Methodology}

In order to rank the Twitter users, we need to define some measure of quality
or usefulness. We will use the in-vocabulary hit rate as our
measure of usefulness since we know from previous experiments that this 
leads to reduction in language model perplexity.
 The task is to predict the utility of a user's data using
information gleaned from that user's social connections. The social connections
are grouped into three categories: mentions (a direct reference to a user
in a public Tweet), followers (people who subscribe to the user's Twitter feed)
and reverse followers (people who the user subscribes to). Our features will
consist of statistics aggregated over each social connection type. For example,
the bigram hit rate statistic will produce three features---namely, the average
bigram hit rate for people who mention, follow, or reverse follow the user. The
statistics that we consider are unigram and bigram hit rates, perplexity,
character set matching and the number of sentences after de-duplication
and filtering. The post-filtering sentence count is a rough measure of how much
data per user ends up actually being used to train the language model. In addition,
we add the degree of each user in the social graph as a feature. This is computed 
separately for each of the three types of edges.

We use random forest regression \cite{randomforest} to predict the user priority
from the features. An alternative ranking scheme, Edge Count Ranking, prioritizes
the download queue based on the cardinality of the set of edges that connect the
seed data to the users in the queue. In this ranking, the next user to be visited
will be the one who has the most social connections to the set of users who have
already been visited. These two ranking methods are compared against a random
baseline that processes the queue in random order.

\subsection{Queue Prioritization Experiment}

We tested our method for prioritizing the crawling queue using Bengali language
Twitter users. In order to properly compare the various methods we first collected
a data set of 10,785 Twitter users. (5,684 users came from our search queries
and an additional 5,103 came from those users' social connections.) The way the
queue ranking works is to first download a small set of users and use that as training
data to learn a ranking for the rest of the download queue. We randomly selected
300 users for our training set. The ranking methods were used to order
the remaining 10,000+ users in the simulated download queue.

\begin{figure}
 \centering
 \includegraphics[scale=0.5]{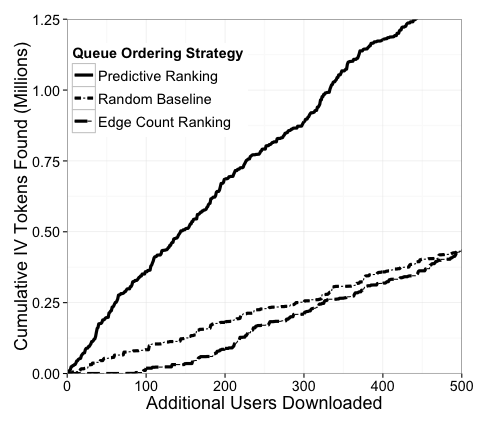}
  \caption{IV Tokens Gathered by Crawling Strategy}
 \label{fig:queue}
\end{figure}

\begin{figure}
 \centering
 \includegraphics[scale=0.5]{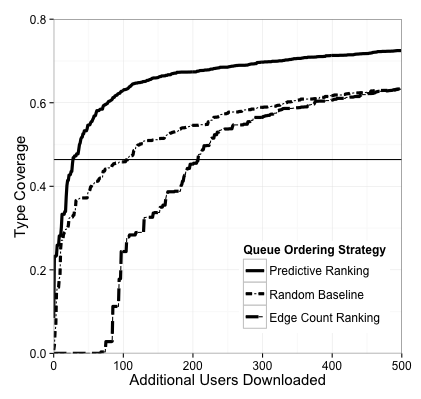}
 \caption{Vocabulary Type Coverage by Crawling Strategy}
 \label{fig:queue_vocab}
\end{figure}

In Figure \ref{fig:queue}, we show the cumulative in-vocabulary tokens obtained as
an additional 500 users are processed from the download queue for each of the three
prioritization strategies.  The predictive ranking provides three times as many 
in-vocabulary tokens as either the edge count ranking or the random baseline. 
Figure \ref{fig:queue_vocab} is similar but shows the cumulative vocabulary coverage (types) 
where the vocabulary is taken from a held-out 
dev set different from the text used to produce the Twitter queries. The horizontal line represents
the type coverage of the in-domain training data.
The predictive ranking covers an additional 10\% of the dev-set vocabulary compared to either of the other strategies.
More than 53\% of the vocabulary types and about 9\% of the tokens in the dev-set are not found
in the in-domain training data.
The Twitter data downloaded using the predictive ranking covers 56\% of these previously unseen
vocabulary types, substantially outperforming the edge count ranking and the random baseline.
The match between the vocabulary of the data obtained with the predictive ranking with the in-domain data
is an indication of the value of the predictive ranking strategy.

The most important features for the predictive ranking (as measured by the average decrease
in node impurity in the random forest regression model) are average number of reverse followers
in the training set, average sentence count after filtering among followers in the training set
and average sentence count after filtering among mentioners in the training set. Even though 
edge features highly influence the predictive ranking model, the edge count ranking is
 worse than the predictive ranking on both of the type coverage and the cumulative IV token metrics. This is
because there are high edge count connections from Bengali speaking Twitter users to international
celebrities, which fools the edge count ranking strategy but not the random forest regression.
(The top followed Twitter user by Bengali speakers is @BarackObama followed by @BillGates.)

We created language models from the downloaded data to compare the effectiveness
of the ranking strategies. The language models used the data
from the 300 users in the training set as well as the additional users selected by each of the
respective ranking strategies. Table \ref{table:crawl} lists the resulting perplexities on
the held-out in-domain data. The predictive ranking LM achieved a
17\% reduction in perplexity over the random baseline after downloading 500 users from the queue.
When less than 500 additional users are downloaded, the predictive ranking has a better perplexity than
the edge ranking. However, once enough data has been acquired the difference in perplexity between
the two ranking systems is small. We reached two conclusions from this experiment: either of the two ranking 
systems is better than not doing any ranking and the predictive ranking has a 
definite advantage at covering the in-domain vocabulary.

\begin{table}[!t]
\centering
\caption{Queue Ranking Strategy LM Perplexity Comparison}
\begin{tabular}{|cccc|} \hline
\multicolumn{1}{|l}{} & \textbf{150 Users} & \textbf{300 Users} & \textbf{500 Users} \\ \hline
Predictive Ranking & 533 & 512 & 490 \\
Edge Ranking  & 547 & 524 & 491 \\ 
Random Baseline & 581 & 572 & 559 \\ \hline
\end{tabular}
\label{table:crawl}
\end{table}

The amount of data used for these language models is small compared to the previous experiments from Table 
\ref{baseline_comparison}.
 As a result, the perplexities are much larger. However, if
more data is downloaded the size of the queue will expand and the predictive
ranking will continue to have an advantage. In a real-world application, the predictive ranking
model would be retrained as additional batches of users are downloaded for maximum performance.

\section{Conclusions and Future Work}
We have shown that Twitter data is useful for modeling conversational speech in a variety of languages.
The high weights assigned to the Twitter data in the mixture models proves the point. As a point of
comparison, using Turkish newspaper text in the mixture model only got 2\% weight compared to the 
27\% weight assigned to the Twitter data.
The combination of Twitter data with the use of class language models is an especially strong result.
The approximately 20\% reduction in perplexity would be hard to achieve by means of collecting additional
in-domain speech data without incurring significant collection costs.

There are still many pieces of unexploited information that could be used to improve the language modeling. 
For example, further analysis and clustering of the Twitter social graph could reveal linguistic subgroups. 
Additionally, selected Tweets have GPS tagging enabled, which could be used to automatically identify and 
model regional dialects.

In the introduction we mention that one reason to prefer collecting data from Twitter is the ease and speed
with which the data can be collected. Our method for prioritizing the download queue builds off this
advantage to increase the amount of data collected and maximize the vocabulary coverage, which justifies
increased effort in searching and crawling. Since the data gathered using the predictive ranking
contains many vocabulary types from held-out data, it could be used for vocabulary expansion. The real-world 
benefit of our predictive ranking approach is sensitive to the rate limits of the Twitter API. For 
simplicity, these considerations were not taken into account in our experiments but, should be addressed
in future work.

\ifCLASSOPTIONcaptionsoff
  \newpage
\fi



%

\bibliographystyle{IEEEtran}

\bibliography{weblm,morefs}

\end{document}

%% file: intro2.tex
\IEEEPARstart{C}{onversational} speech is very different in style from broadcast news and prepared speeches, so the language models used in automatic speech recognition (ASR) of conversational speech rely on training from speech transcripts, which are more costly to obtain than more formal written text.  Thus,
data sparsity is typically the limiting factor for language model performance. For many less well-studied languages, there is little transcribed speech available and obtaining additional training data in the target domain is no easy task. An attractive alternative to collecting additional data is to direct that effort towards building models that require less training data. Since little additional work is needed to reuse these models, the benefit of this effort is compounded with each new language thus lowering the barrier for 
bringing ASR to low-resource languages. While new modeling approaches are valuable, it is the case that even these models benefit from additional data, and
researchers have long been exploring mechanisms for using out-of-domain data in combination with a small in-domain corpus to build more robust language models. Different sources of spontaneous speech can help with common words, but these are often not available for less well-studied languages.  Even with such data, covering domain-specific vocabulary typically means using written text sources for training language models for ASR. Finding informal text that is useful for modeling conversational speech can be a challenge. 

The Internet has been an attractive place to go for researchers looking to expand
 their training data, as described in the next section. 
However, if done without care, pulling large amounts of data from the Internet will give no benefit. For example, in work on recognizing English conversational speech in broadcast talk shows, the Google n-grams provided no benefit to perplexity or word error rate \cite{Marin09}.  Here, we instead propose to use only Twitter for collecting out-of-domain data for ASR language models for conversational speech.

We have several reasons to prefer searching Twitter
for data instead of the whole Internet. First, Twitter has a semi-structured format that makes
it easy to select the user generated content and ignore ads, boilerplate text and other
unwanted text. Second, the writing style is somewhat conversational which is a better match
for the conversational speech domain than other types of writing often found on the Internet. Third, since
Twitter is used all over the world, we will likely be able to find good data for most of the
languages that we care about. Lastly, Twitter provides a friendly API
that makes it easy to access a lot of data quickly. When working under a time constraint, we were
able to collect as much as 6 million sentences from low resource languages in under 24 hours. 

Of course, Twitter text is notoriously noisy in that tweets often contain abbreviated or misspelled forms of words, as well as URLs, hashtags, usernames, etc.  In this work, we describe simple language-independent methods for handling these that require filtering out a large amount of data but lead to surprisingly useful text.  In fact, we find that Twitter text is more useful for learning word classes than the original in-domain text. The word classes can be used to achieve large reductions in perplexity
for low-resource language models. Further, we introduce a prioritizing scheme for downloading text from ``useful'' users that results in a significant improvement in vocabulary coverage over random selection.

In the remainder of the paper, we begin by summarizing prior work on leveraging text harvested from the web in language modeling in Section \ref{section:related}, to provide the context for our work. Next, in Section \ref{section:twitter}, we describe our initial process for collecting data from
Twitter and two methods for using the data in language modeling. Section \ref{section:experiments}
contains details of experiments demonstrating the effectiveness of the collection and modeling methods.
Finally, in Section \ref{section:crawling} we describe a method for prioritizing the crawling queue
and how it can be applied to improve the utility of additional collected data.

%% file: related2.tex
This work brings together two strategies for improving language models for conversational speech in limited training data scenarios -- web text collection and class language models -- but changes key elements in the particular approach for both.

As mentioned previously, there have been several efforts aimed at using text gathered from the web. Early work used the number of search results returned by the search engine as a proxy for the n-gram probability \cite{Zhu_Rosenfeld:2001}, but this method does not capture the domain characteristics of the target task. The penalty for domain-mismatch is quite severe. When working with conversational speech, a small set of conversation transcripts makes for better training data than 100 times as much newswire text \cite{Rosenfeld00}. Other studies find that even speech transcripts can be of little utility if there is a formality mismatch, e.g.\ using broadcast news as out-of-domain data for a conversational task.

A series of papers used frequent in-domain n-grams as queries to Google for collecting text \cite{Bulyko_etal:2003,Schwarm+04,Bulyko07} and perplexity filtering for improving the match to the target domain. Good results were obtained on a variety of conversational speech recognition tasks by several sites. Variations on this approach changing the query generation and filtering stages have also been explored \cite{Sethy+05,Sarikaya+05,Yoshino13} or by using cross-entropy rather than perplexity as a data selection criterion \cite{MooreLewis10,Axelrod11}. Recent work has shown reduced perplexity and word error rate for a task of transcribing university YouTube videos \cite{Lecorve+12}. Unfortunately, this method was not effective for the low-resource languages we are working with, perhaps because of the specific choice of languages. Other recent alternatives have used targeted text collections, e.g. broadcast news transcripts \cite{Marin09} and RSS feeds \cite{web20}. 

Schlippe et al. \cite{web20} also collected Twitter text for augmenting the language model training data and the vocabulary for a French news transcription task, which led to a small (1.5\% relative) reduction in word error rate.  The tweets are collected using topic words from the RSS feeds for a 5-day period before the show, and text normalization uses language-specific knowledge (dictionary spell checking and abbreviation expansion). Key differences in our use of Twitter text are the focus on general conversational speech rather than recent news topics, simple text normalization methods that are minimally language-dependent, and the introduction of a new priortization strategy for rapid collection.

A standard approach to incorporating web text in language modeling is via n-gram mixture models, where component n-grams are trained on different data sources.
Class language models \cite{brown92} have been used to reduce the number of parameters in the model for combining data from different domains \cite{IyerOst97}, but also to provide more degrees of freedom in mixture model combinations of data from different sources by allowing history-dependent mixture weights using classes \cite{Bulyko07}. Because of the limited resources, the work here emphasizes reducing the number of parameters, but the novel contribution is using the out-of-domain data to learn the word classes, which turn out to be more powerful than those learned on the sparse in-domain data.
